\documentclass[10pt,letterpaper,conference]{IEEEtran}

\usepackage{cite}
\usepackage{amsmath,amssymb,amsfonts}
\usepackage{graphicx}
\usepackage{textcomp}
\usepackage{xcolor}
\usepackage{caption}
\usepackage{subcaption}
\usepackage{comment}
\usepackage{paralist}
\usepackage{tikz}
\usepackage{url}
\usepackage[hidelinks]{hyperref}

\IEEEoverridecommandlockouts
\def\BibTeX{{\rm B\kern-.05em{\sc i\kern-.025em b}\kern-.08em
    T\kern-.1667em\lower.7ex\hbox{E}\kern-.125emX}}

\newcommand\copyrighttext{%
\footnotesize This is the preprint accepted for publication in the 6th International Workshop on Wireless Sensors and Drones in the Internet of Things (Wi-DroIT), held in conjunction with the 20th International Conference on Distributed Computing in Smart Systems and the Internet of Things (DCOSS-IoT), Abu Dhabi, United Arab Emirates, 29 April - 1 May 2024. This version is released under a CC-BY license according to the requirements of the Horizon Europe programme that has provided funding for this work. The final version is available at \href{https://doi.org/10.1109/DCOSS-IoT61029.2024.00058}{https://doi.org/10.1109/DCOSS-IoT61029.2024.00058}.}
\newcommand\copyrightnotice{%
\begin{tikzpicture}[remember picture,overlay]
\node[anchor=north,yshift=-10pt] at (current page.north) {\fbox{\parbox{\dimexpr\textwidth-\fboxsep-\fboxrule\relax}{\copyrighttext}}};
\end{tikzpicture}%
}

\begin{document}

\title{Should I Stay or Should I Go: A Learning Approach for Drone-based  Sensing Applications   
}

\author{\IEEEauthorblockN{
Giorgos Polychronis, Manos Koutsoubelias and Spyros Lalis}
\IEEEauthorblockA{\textit{Department of Electrical and Computer Engineering} \\
\textit{University of Thessaly}\\
Volos, Greece \\
\{gpolychronis, emkouts, lalis\}@uth.gr}
}

\maketitle

\copyrightnotice

\begin{abstract}
Multicopter drones are becoming a key platform in several application domains, enabling precise on-the-spot sensing and/or actuation. We focus on the case where the drone must process the sensor data in order to decide, depending on the outcome, whether it needs to perform some additional action, e.g., more accurate sensing or some form of actuation. On the one hand, waiting for the computation to complete may waste time, if it turns out that no further action is needed. On the other hand, if the drone starts moving toward the next point of interest before the computation ends, it may need to return back to the previous point, if some action needs to be taken. In this paper, we propose a learning approach that enables the drone to take informed decisions about whether to wait for the result of the computation (or not), based on past experience gathered from previous missions. Through an extensive evaluation, we show that the proposed approach, when properly configured, outperforms several static policies, up to $25.8\%$, over a wide variety of different scenarios where the probability of some action being required at a given point of interest remains stable as well as for scenarios where this probability varies in time.    
\end{abstract}


\maketitle

\section{Introduction}
\label{sec:intro}


Unmanned aerial vehicles (drones) are being used in an increasing number of applications and will become a key component of next-generation smart infrastructures designed to support autonomous sensing and actuation. 
Polycopter drones, being able to quickly move to different points of interest and to hover right on top of the respective location to gather data or to perform certain actuation from a specified altitude, are particularly  suitable for this role. 
However, due to the limited autonomy of such drones, it is typically desired to minimize the time it takes 
to visit the points of interest. 

Notably, some 
missions can be data-driven, meaning that the drone may need to perform further action depending on the data collected via 
its onboard sensors. In this case, the data must be processed in order to decide whether further action needs to be taken. For example, in a smart agriculture scenario, if pest is detected at a specific location, the drone can immediately spray this specific location with pesticide. 
Such actions must be performed on the spot, while maximizing the area / number of locations that can be scanned before the drone 
is forced to return home to change batteries. 

One way to reduce the mission time is optimally plan the drone's path so that it visits all points of interest in the shortest possible time. Another way is to accelerate the processing of the sensed data by offloading the computation to the cloud or nearby edge servers, thereby reducing the time spent by the drone at each point of interest before moving to the next one. Both aspects are well studied in the literature. 

In this paper, we explore a complementary approach for reducing the time of such missions. Namely, instead of waiting until the computation finishes, we allow the drone to move to the next point of interest right after sensing. This way, time is gained if no action needs to be taken at the previous point of interest. 
Else, the drone must spend time to return to the previous point of interest to perform the required action. 
To tackle the problem, we propose a decision algorithm that learns from the drone's past experience, 
thereby allowing the drone to more informed decisions in future missions. 

The main contributions of the paper are: 
(i) We provide a formal  
formulation  
for the above decision problem. 
(ii) We propose an autonomous, learning approach to tackle the problem, which uses experience from previous missions. 
(iii)  We evaluate the proposed approach via extensive simulation experiments, for non-trivial scenarios where the probability of action varies in both space and time. 
(iv) We validate the 
accuracy of the simulator used to perform the  
experiments against the real flight behavior of a quadcopter drone and software-in-the-loop mission execution of a real autopilot.  We evaluate the approach in the field using a real drone.

The rest of the paper is structured as follows.
In Section~\ref{sec:rel} we give an overview of related work. Section~\ref{sec:prob} formulates the problem, and Section~\ref{sec:approach} describes the proposed approach. Section~\ref{sec:validation} presents the validation of the simulator used for our experiments, while Section~\ref{sec:eval} evaluates the proposed approach. Finally, Section~\ref{sec:conc} concludes the paper.

\section{Related Work}
\label{sec:rel}

The problem we tackle in this paper has similarities with other problems that have been investigated in the wider domain of autonomous vehicles and drones. Below we provide a brief overview and highlight the differences with our work.

Some works focus on reducing the time it takes to perform heavyweight computations on autonomous vehicles, by offloading such computations to the cloud or nearby edge infrastructure. For instance, in \cite{messous2020edge} drones perform a 3D mapping in an unknown area, taking offloading decisions for the required computation tasks. In~\cite{chen2020intelligent}, multiple drones offload their tasks to the edge. The authors propose an offloading algorithm that predicts the channel quality based on historical data. In~\cite{icfecKasidakis} the drone employs a runtime heuristic to decide whether it is worth offloading the computation to an edge server instead of performing it locally. 
This decision is based on previous experience about the end-to-end response time of each server. 
Our work is complementary to these efforts, as it focuses on minimizing the mission time of the drone by taking smart decisions about whether to actually wait for such computations to complete before moving to the next waypoint.
\cite{bandarupalli2023vega} 
presents a drone-based 
system for detecting ground targets. One of the proposed strategies is to fly at high altitude covering a wider area and, if a target is detected, instruct the drone to 
perform a close-up sensing. This is similar to the application pattern assumed in our work. However, the authors focus in the tradeoff between detection delay, coverage and detection quality, while we reduce the mission time by allowing the drone to move to the next point of interest before the results of the previous detection attempt become available.

A major component of total mission execution time in  
vehicle-related applications comes from travelling between the different points of interest. Thus, many works focus on the problem of path planning
~\cite{gugan2023path}. Also, a lot of work has been done in different vehicle routing problems, especially for electrical vehicles which have limited autonomy and must recharge or change batteries~\cite{kucukoglu2021electric}. 
Other works use drones as a mobile edge server or base station, providing resources to nodes on the ground. For example, \cite{zhang2018stochastic} uses a drone to serve mobile devices, 
optimizing the drone's trajectory based on the stochastic task arrival times. In~\cite{asim2022evolutionary}, multiple drones are used to serve users, planning the drone trajectories so as to minimize the total energy consumption. \cite{lin2022multiobjective} presents similar work for UAVs serving IoT devices. 
In our work, we take the (potentially optimized) drone path as input, and try to reduce the mission time by overlapping computation with flying.

The problem we study 
has some similarities with the vehicle routing problems (VRP) with stochastic aspects~\cite{ritzinger2016survey}. One such case is having a vehicle with limited supply capacity visiting customers in different locations to serve their stochastic demands which are not known in advance. 
If it turns out that the demand can not be satisfied, the vehicle must take some corrective action, e.g., return to a depot to resupply and return to the customer~\cite{christiansen2007branch, goodson2012cyclic,  gonzalez2018simheuristic}. The cost of such potential detours is taken into account when planning the path of the vehicle. In our work, a similar corrective action is taken when the drone decides to go to the next waypoint without waiting for the computation to complete, but (after the computation ends) it turns out that further action is needed at the previous point of interest. The proposed decision approach takes into account the potential gains and penalty, weighted with the respective estimated probability, to minimize the mission time. 
Another key difference is that, in our case, the potential gain or penalty is affected not only by the probability of having to perform an additional action at the point of interest, but also (and quite substantially so) by the processing delay.

Less related to our  
application domain, a large body of literature has studied the problem of load estimation 
in cloud computing systems~\cite{VMsched1,VMsched2,VMsched3,VMsched4}. The main issue at hand is to estimate the resources 
that will be required to serve a given job or application VM 
in order to take more informed job scheduling decisions, e.g., regarding the node that will be used to run a new job, the migration of existing jobs on different nodes, and the packing of multiple jobs on fewer nodes (consolidation) so that other nodes can be turned-off to save energy. Similar to our work, this estimation is based on previous experience, typically using a form of regression over the loads recorded in previous scheduling periods. In addition, we study increased system dynamics scenarios, where the previously recorded resource usage becomes partly or even completely invalid and the decision making algorithm must strongly discount or even forget previous experience.

\section{Problem Formulation}
\label{sec:prob}

\subsection{Missions}

We consider data-driven missions where the drone visits certain points of interest to perform some sensing, process the sensed data, and maybe -- depending on the outcome of the computation -- perform an additional action. 

We assume that the path 
of the drone is pre-computed using some planning algorithm (not the focus of this work). Let this be encoded as sequence of waypoints in the order that they must be visited by the drone,  
$[wp_1, wp_2, ..., wp_N]$. 
The first and last waypoint $wp_1$, $wp_N$ is the home location from where the drone starts and returns, 
while $wp_i, 2 \leq i \leq N-1$ correspond to the points of interest. 

Let $senseT$ be the time that is needed for the drone to perform the required sensing at each such waypoint, and $procT$ the time that is needed to process the sensed data.
Without loss of generality, we assume that data processing 
generates a so-called detection event, only if a certain object or situation is detected  
requiring further action from the drone. 
Note that such event may not be generated, but the drone does not know this before the computation is finished. Let $e_i$ encode whether a detection event was generated for $wp_i$ or not, taking values $1$ and $0$, respectively. 
Let $actionT$ be the time  
needed for the drone  
to perform the additional action 
if $e_i = 1$.

\subsection{Flight model}

We focus on polycopter drones, which can 
takeoff and land vertically,  
and 
can hover  
above a specified location, if required. We assume an obstacle-free mission area, where the drone can move in  
straight lines. 
Let  $flightT(wp_i,wp_{i+1})$ be the time it takes for the drone to fly from $wp_i$ to $wp_{i+1}$. Similarly, let $takeoffT$ be the time needed to perform a vertical takeoff, and $landT$ be the time needed for a vertical landing. 

In principle, a relatively good estimate of these delays 
can be calculated analytically 
based on the distance to be covered, the drone's vertical and horizontal acceleration/deceleration etc.
In our work, to have more accurate results, 
we use the take-off, flight and landing times of a realistic software-in-the-loop (SITL) autopilot configuration  
(see Section~\ref{sec:validation}).

\subsection{Decision, time gain and time penalty}

After sensing at waypoint $wp_i$, the drone must decide whether it will keep hovering at that location waiting for the completion of the data processing computation, or move on to the next waypoint $wp_{i+1}$. Let this decision be encoded as $d_i$, taking values $0$ and $1$, respectively. 

We take as a ``baseline'' the decision to wait at $wp_i$ ($d_i = 0$). Then, the drone will take $senseT + procT$ to perform the required sensing and wait for the computation to finish. It will also take $flightT(wp_i,wp_{i+1})$ to move to the next waypoint.

We now focus on the case where the drone decides to move from $wp_i$ to $wp_{i+1}$ ($d_i = 1$) as soon as sensing is completed. If no detection event is generated as a result of data processing, it will save $procT$ vs having waited at $wp_i$ for the computation to finish. 
If, however, a detection event is generated, it will waste time vs having simply waited at $wp_i$. This is because the drone must stop moving towards $wp_{i+1}$ and start moving backwards to approach $wp_i$ so that it can perform the required action. Let the time penalty for this be  $penaltyT(wp_i,wp_{i+1},procT)$. 

Again, rather than estimating this penalty in an analytical way, this is measured using 
the SITL autopilot, as a function of $procT$. For simplicity, we assume that the processing time is smaller than the flight time between any two waypoints, $procT < flyingT(wp_i,wp_{i+1})$. As a consequence, the drone always discovers whether it needs to return to $wp_i$ before having reached $wp_{i+1}$. It is  possible to generalize our approach to handle the special case where the drone returns to the previous point of interest after having reached the next one, but this is out of scope of this paper. 

\subsection{Mission time}

Based on the above, the mission time can be calculated as:

\begin{equation}
\label{eq:missionT}
\begin{split}
    missionT = takeoffT + flightT(wp_1,wp_2)+ \\ \sum_{i=2}^{N-1}(visitT_i) +  landT
\end{split}
\end{equation}

\begin{equation} 
\label{eq:visiTi}
\begin{split}
visitT_i = & senseT + procT + flightT(wp_i,wp_{i+1}) \\
& + e_i \times d_i \times penaltyT(wp_i,wp_{i+1},procT) \\
& - (1-e_i) \times d_i \times procT \\
& + e_i \times actionT
\end{split}
\end{equation}

\noindent where $visitT_i$ is total the time spent by the drone to complete the visit of $wp_i$. 
This includes the ``baseline'' time,  
adding the penalty when the drone wrongfully decides to move to the next waypoint before the computation finishes, and, conversely, subtracting the gain when the drone correctly decides to  
move on. It also includes the time for performing a follow-up action if needed (when a detection event is generated).   

Obviously, the goal is for the drone to take decisions $d_i$ so as to minimize $visitT$ and consequently also $missionT$. The problem is that the  generation of the detection events $e_i$ is unknown to the mission planner and thus cannot be encoded in the logic of the mission program that controls the drone.

\section{Approach}
\label{sec:approach}

We assume that the generation of detection events $e_i$ at each waypoint $wp_i$ at time $t$ is governed by an underlying probability $p_{wp_i,t}$. In other words, we assume that the probability of needing to take further action is both location and time dependent. 
If the drone knew this probability, it would be able to take an informed decision at each $wp_i$ as follows:

\begin{equation}
\label{eq:bestdecision}
    d_i \gets \arg\max_{c=0,1} E(c,wp_i,t)
\end{equation}

\begin{equation}
\label{eq:wait}
E(0,wp_i,t) = p_{wp_i,t} \times penalty(wp_i,wp_{i+1},procT)
\end{equation}

\begin{equation}
\label{eq:go}
E(1,wp_i,t) = (1-p_{wp_i,t}) \times procT 
\end{equation}

where $E(c,wp_i,t)$ is the expected time saved at $wp_i$ and time $t$ by choosing to take decision $c$.

In the following, we present an approach for the drone to autonomously learn the detection event probability $p_{wp_i,t}$. 

\subsection{Learning from past experience}
\label{sec:approach_knowledge}

During the execution of a mission, the drone records whether a detection event was generated for each waypoint at the time when it performed the sensing. 
This information is stored in the form of separate entries 
$\langle wp_i,t,e_i \rangle$, where $t$ is the time of day (in the granularity of hours) the sensing was performed at waypoint $wp_i$ and $e_i$ denotes whether data processing lead to a detection event that requires further action from the drone. The entries for each $wp_i$ and $t$ are kept in an ordered list according to the time when these were created.

Before starting a new mission, a regression algorithm is run over the experience dataset to estimate the event generation probabilities $p_{wp_i,t}$ for the upcoming mission. 
Only the experience that is actually relevant for the mission at hand is taken into account. More specifically, assuming the mission starts at $t_{start}$ and the drone has an operational autonomy of $opT$, we only consider the entries $\langle wp_i,t,e_i \rangle$ where $wp_i$ is among the points of interest to be visited by the drone and $t$ falls inside the time period $[t_{start}..t_{start}+opT]$.

Our design is modular making it possible to employ different regression algorithms in a plug-and-play manner. Specifically, we run experiments using linear regression~\cite{linear}, decision tree regression~\cite{tree} and  Bayesian  regression~\cite{bayesian} (see Section~\ref{sec:eval}).
Notably, we do not make any 
assumptions about the potentially spatial or time dependence of $p_{wp_i,t}$. We simply pass to the regression algorithms(s) all the data from the experience memory that matches the context of the mission  
(points of interest and time period of mission execution). 

\subsection{Experience memory size and reset}
\label{sec:memory}

In the general case, it is not possible to  
assume infinite experience memory. Also, if the detection event probability changes in time, past experience becomes less useful. 
For this reason 
we only keep the $H$ most recent entries for each $wp_i$ and $t$. We refer to $H$ as the size of the experience memory. 
In the evaluation, we experiment with different 
values for $H$. 

However, 
if $p_{wp_i,t}$  
changes significantly,  
even recent 
experience may be 
quite misleading. To handle such scenarios, we  
introduce an anomaly detection mechanism 
which triggers a reset of the experience history (drop all $\langle wp_i,t,e_i \rangle$ entries except the ones corresponding to the recent mission) whenever an anomaly is detected.  
More specifically, 
during mission execution, we record the time penalties the drone has experienced due to wrong decisions, 
which are aggregated to a total penalty on a daily basis.
An anomaly is signalled if the penalty for the current day becomes greater than the maximum daily penalty that was experienced over the  
first $H$ days of learning, 

As an alternative option, 
we use the isolation forest method that has been proposed in the literature for anomaly detection~\cite{isoforest}. We run this method on the 
penalties the drone has experienced in the last $H$ days with contamination (outlier proportion to the total data) set to $1\%$.

\subsection{Drone control logic}

Before starting the mission, the regression algorithm is used to estimate the probabilities of event generation at the relevant waypoints and times of day. Then the drone takes-off and starts visiting the waypoints according to the specified mission plan, taking each time a decision based on 
the proposed learning approach; if the decision is to ``go'' but this turns out to be wrong, the drone is instructed to return to the previous point of interest to perform the  
required action before moving to the next one.  
When the last point of interest is visited, the drone returns to the home location, lands and saves the experience of the mission. 
Finally, if an anomaly is detected  
the experience memory (to be used in the next mission) is reset.

\section{Validation of High-level Simulator}
\label{sec:validation}

In this section, we validate the high-level simulator 
used for the extensive evaluation of the proposed approach. 
As a first step, we run a  
mission in the field using a real drone, and compare the observed flight behavior vs  
running the same mission with the autopilot in software-in-the-loop (SITL) mode.  
Then, we compare the results we get for a more complex mission scenario when using the SITL autopilot vs the high-level simulator. We show that the differences  
are  
small 
thus the high-level simulator can 
serve as a practical evaluation tool without having to  
do real field tests. 

\begin{figure}
     \centering
     \begin{subfigure}{0.4\linewidth}
         \centering
         \includegraphics[width=\linewidth]{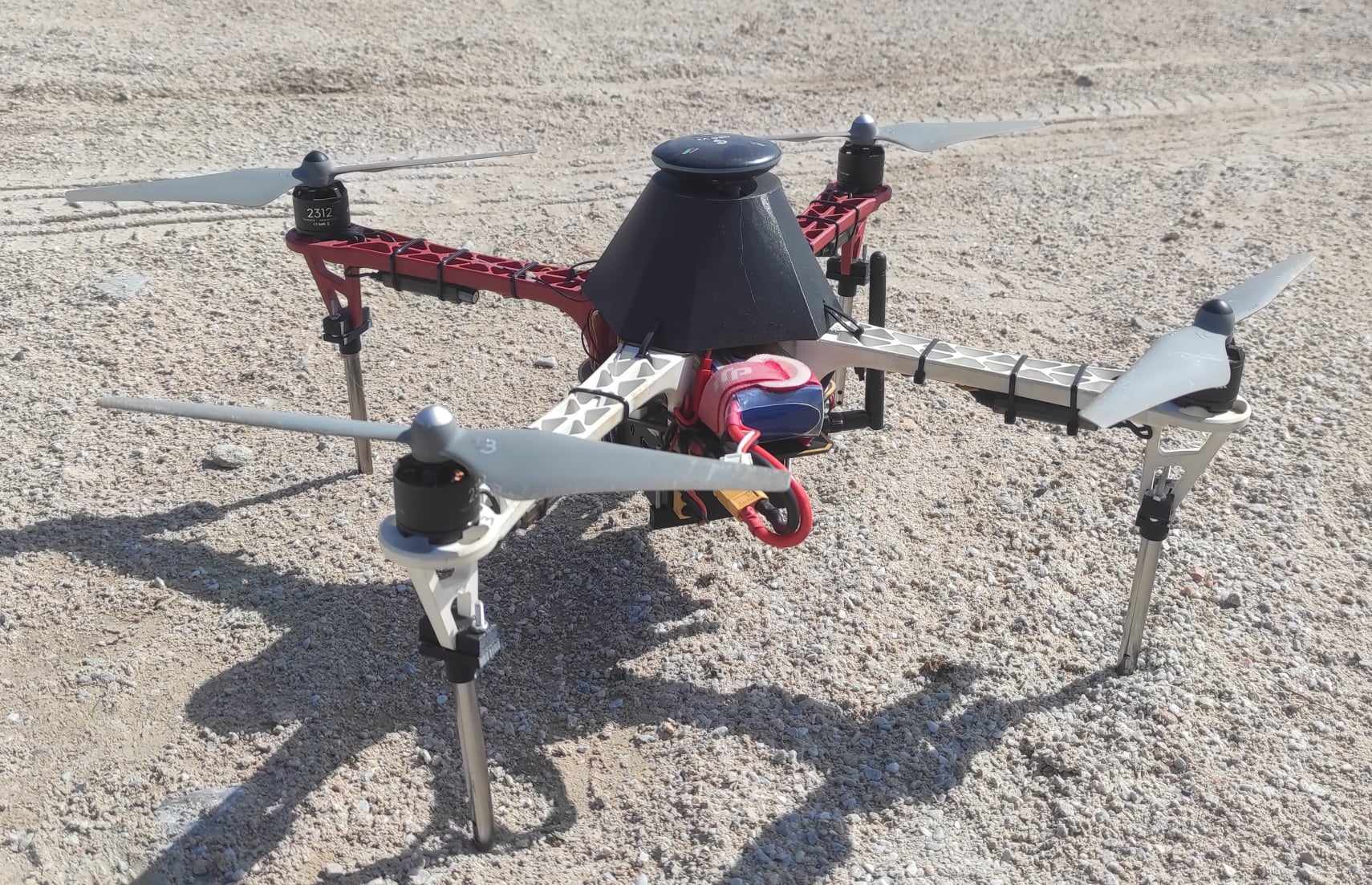}
         \caption{Custom drone.}
         \label{fig:drone}
     \end{subfigure}
     \begin{subfigure}{0.55\linewidth}
         \centering
         \includegraphics[width=\linewidth]{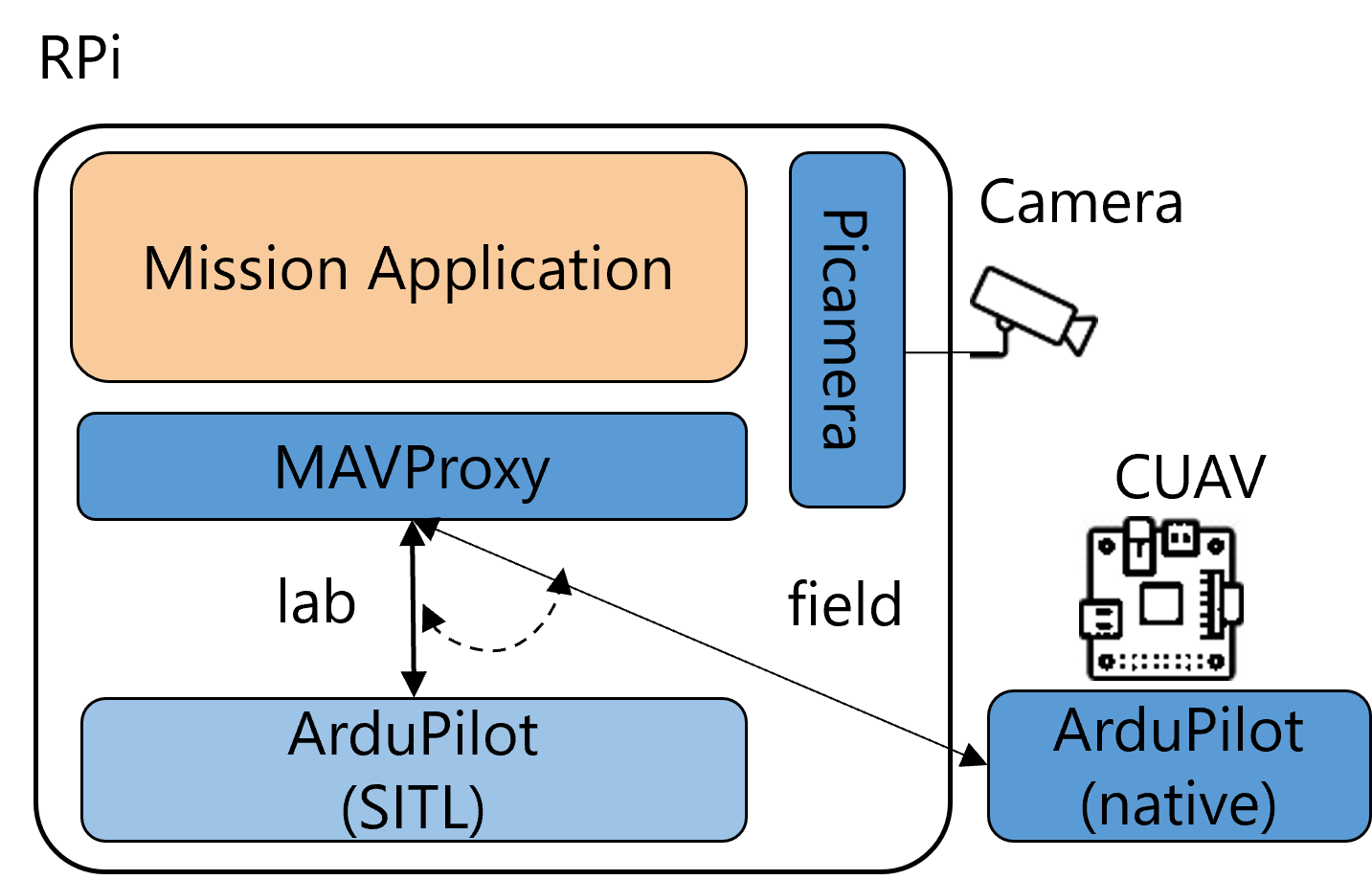}
         \caption{Autopilot in field vs lab setup.}
         \label{fig:swconfig}
     \end{subfigure}
     \caption{Drone in the field vs SITL configuration.}
     \label{fig:field}
\end{figure}

\begin{figure*}
  \includegraphics[width=\textwidth,height=4cm]{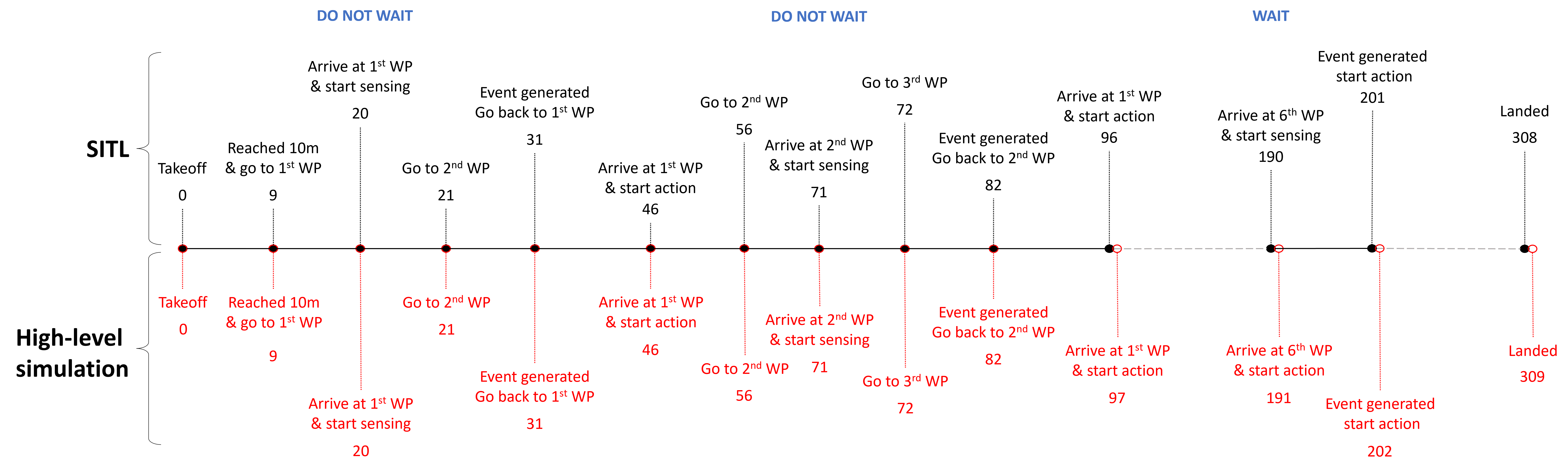}
  \caption{Timeline of a complex  mission execution scenario using the autopilot in SITL mode vs the high-level simulator.}
  \label{fig:timeline}
\end{figure*}

\subsection{Validation of SITL vs real flight behavior}

For the field experiments, we use a custom quadcopter drone, shown in Figure~\ref{fig:drone}. This runs the ArduPilot autopilot~\cite{ardupilot} on a CUAV Nano v5 board~\cite{cuav}. The drone is also equipped with a Raspberry Pi (RPi) with a camera, serving as an indicative on-board sensing and computing platform. The RPi is connected to the autopilot board via serial and hosts the application code, which runs the desired mission by retrieving the status and sending commands to the autopilot using the MAVLink protocol~\cite{mavlink} through the MavProxy library~\cite{mavproxy}.

For practical reasons we perform our experiments in an empty field 
where the drone moves back-and-forth between two waypoints (WP1 and WP2) $50$m apart, representing points of interest. 
At WP1, the drone performs the required sensing and immediately starts moving towards WP2. 
We assume  
that processing always raises a detection event, 
so the drone stops  
and returns to WP1. In contrast, when the drone visits WP2 it always waits for the computation to finish before moving to WP1. All movements between the two waypoints are performed in a straight line at a target cruising speed of $4$m/s. 
Despite 
its simplicity, this scenario captures the basic flight behavior of the drone that is relevant for our study.

The RPi takes about $1$ second to capture an image and from 8 to 12 seconds to perform the computation for object detection via yolov8x~\cite{yolo} for a 1080p image downscaled to 480p and 640p, respectively. We have chosen  
at least 480p as the image input size and the slowest but most accurate yolov8x model to allow objects to be detected with 0.95 confidence. 
In the above  
experiments, the RPi sleeps instead of  
performing the computation as we do not rely on the results. We perform different tests for different (virtual) computation times of $8$, $10$ and $12$ seconds. 
For each scenario, we let the drone repeat these waypoint visits $10$ consecutive times, and record the total mission delay  
as well as the  
time  
it takes for the drone to stop moving towards WP2 and return to WP1.

We 
repeat these  
experiments with the autopilot running in the  
SITL configuration~\cite{SITL} coupled with a physics model that simulates the dynamics of the aerial vehicle. The respective software configuration is shown in Figure~\ref{fig:swconfig}. Note that the 
mission code remains exactly the same as in the field tests 
except that it interacts with the SITL autopilot configuration instead of the autopilot board. 
Overall, the results of the SITL experiments are very close to those of the field experiments,  
with an average deviation of the mission time about 
$10\%$. 
This confirms that the SITL configuration can reproduce the drone behavior with sufficient accuracy so that it can be used as a tool for experimenting with more complex scenarios. 

\subsection{Validation of high-level simulator vs SITL}

As an alternative to the accurate SITL-based mission execution, we have developed a lightweight high-level simulator that captures all the delays of the drone's movements that are relevant for our study, based on the SITL measurements, without using the autopilot or an underlying physics model.

To validate the high-level simulator vs the SITL-based mission execution, we use a more complex mission. More specifically, the drone visits $8$ waypoints, in half of which the drone decides to wait for the computation to finish while for the other half it decides to go to the next waypoint. At each waypoint, we set the generation of the detection event so that half of the drone's decisions are correct and the other half are wrong. In other words, there are $2$ cases where the drone wrongfully decides to move to the next waypoint before the computation finishes and must return back to the previous waypoint. Also, in $2$ cases the drone takes the wrong decision to wait for the computation even though this does not generate a detection event. In this experiment, we set the computation time to $10$ seconds. As in the field tests, the target cruising speed of the drone is set to $4$m/s. 

We run the same mission using the SITL configuration and the high-level simulator, recording all relevant events, including time to take-off and land, flight time between waypoints, time spent at each waypoint and the time penalty when deciding to go but then need to return back to the previous waypoint. 
Figure~\ref{fig:timeline} shows the event timeline for each of these executions. The black events correspond to the execution of the mission using the SITL configuration, while the red events correspond to the mission execution using our high-level simulator.
The time difference for the same events is in the order of a few milliseconds, 
leading to  
an aggregate deviation of merely a couple of seconds for the 
entire mission.

We have performed several variations of the above experiment, confirming that the high-level simulator faithfully reproduces the results of the SITL-based mission execution. However, 
the latter is far more time-consuming (it takes the same time as if the drone were actually flying in the real world), whereas the high-level simulator takes just a few seconds. Therefore, we conduct our extensive evaluation 
using  
the high-level simulator. As an extra precaution, we have  
run selected scenarios from those experiments via SITL, which always produced practically identical results. 

\section{Evaluation}
\label{sec:eval}

\subsection{Path plan, detection probabilities, mission scenarios}
\label{sec:baseplan}

\begin{figure}
    \centering
    \includegraphics[width=0.85\linewidth]{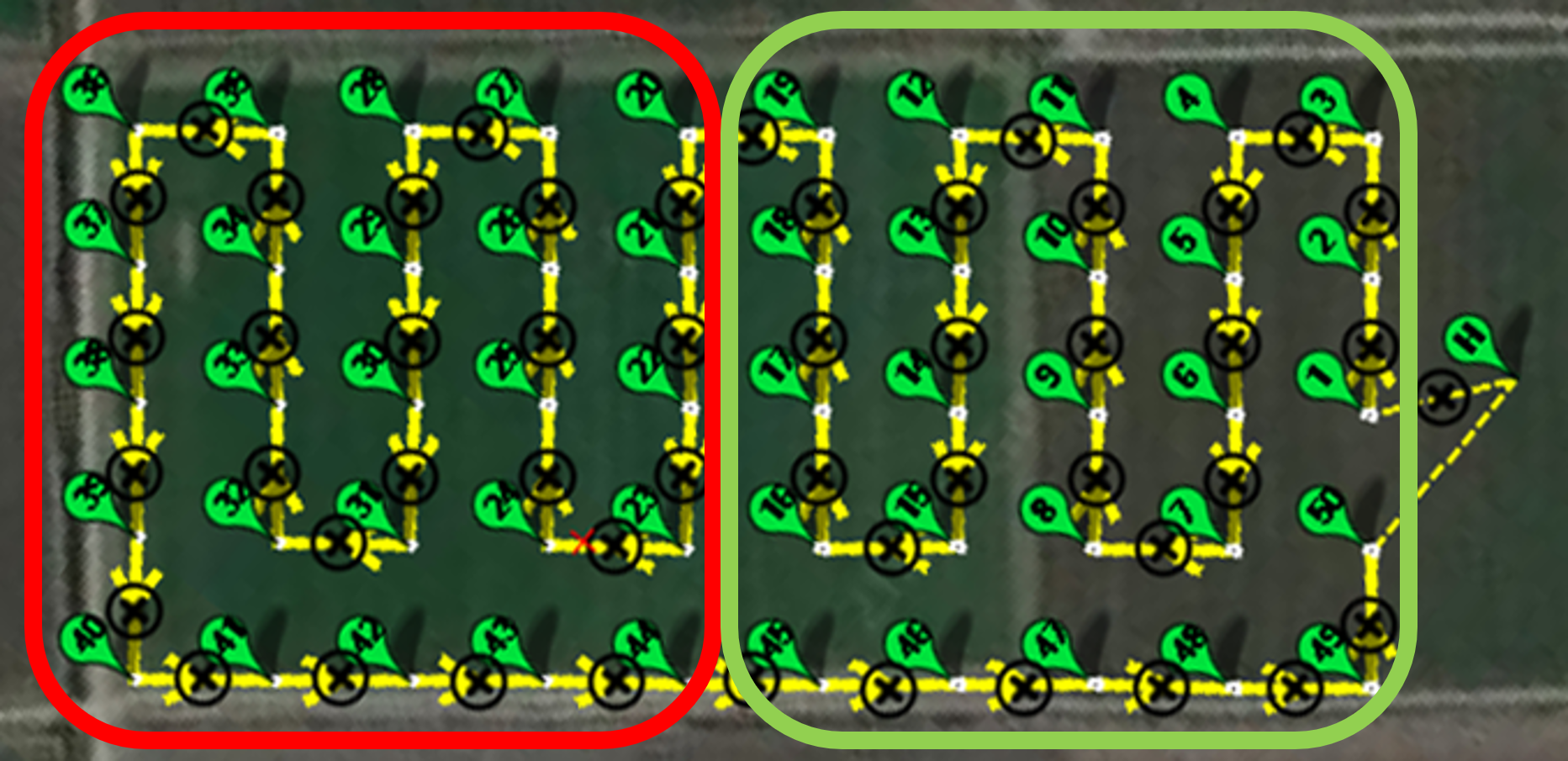}
    \caption{Mission area and path plan.}
    \label{fig:mission}
\end{figure}

For our evaluation we use a  
$450$m x $200$m area of interest, consisting of 50 waypoints spaced $50$m apart in grid-like manner, as shown in Figure~\ref{fig:mission}. In all experiments, the drone starts from a nearby  
location, visits these waypoints following  
a pre-specified path plan, and returns to home (yellow line).   

Learning wise, it is more interesting to have
certain points of interest produce detection events with different probability, also as a function of time. 
To model this case, we separate the mission area in two regions denoted by the colored rectangles in Figure~\ref{fig:mission} with different detection probabilities, as a function of daytime. 
More specifically, waypoints in the green region have event detection probability $0.1$ throughout the entire day except the ``time window'' 12:00-15:59 where the probability increases to $0.6$. Conversely, for the waypoints in the red area, the event detection probability during the entire day is $0.6$ dropping to $0.1$ during the above time window. 

\begin{figure*}
    \centering
    \includegraphics[width=0.85\linewidth]{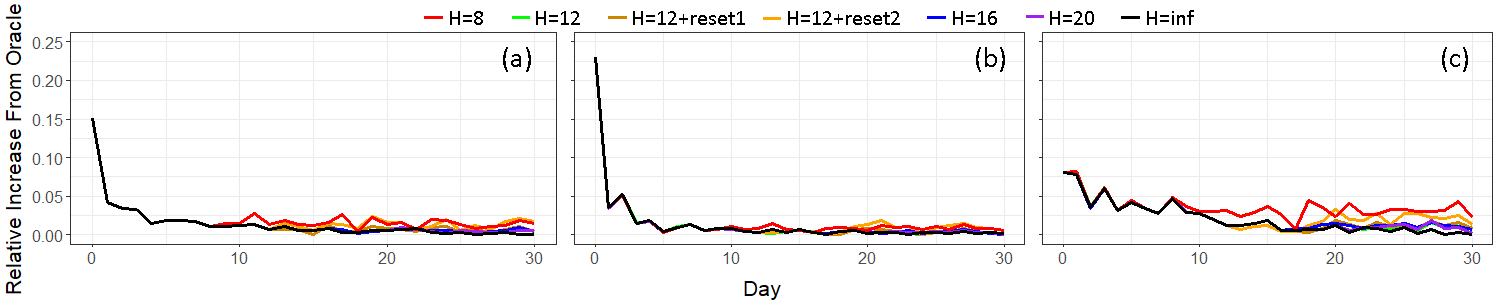}
    \caption{Experience memory comparison. Stable world. (a) 
    out scenario, (b)  
    out-in scenario, (c)  
    in-out scenario.}
    \label{fig:past_fixed}
\end{figure*}

Note that the drone first scans most of the green region, then scans the entire red region, and finally scans the remaining part of the green region.
The start time of a mission  
affects the drone arrival time at the points of interest  
hence also the probability for generating detection events at those locations. 
For this reason, we experiment with three  
scenarios where (i)~the mission takes place outside the time window (out), (ii)~the mission starts outside the time window and ends inside it (out-in), and (iii)~vice versa (in-out).  
In the first scenario, half of the mission waypoints have event detection probability $0.1$ and the other half $0.6$. In the last two scenarios, we set the  
start time so that approximately half of the mission  
time (when taking optimal decisions based on the true underlying event detection probability)  
is spent outside and half inside the time window. 
Thus, in the  
out-in scenario, most visited waypoints have event detection probability $0.1$ whereas in the  
in-out scenario most have $0.6$.
Note that the two mission scenarios are not entirely symmetrical as 
they start 
with a different offset from the respective time window boundary. 

While the 
above captures variation 
in both space and time, seen from a higher perspective, this  
corresponds to a \textit{stable world} as this pattern  
applies to all days. However, in the general case, the world may change, perhaps even radically. We explore this  
via the \textit{changing world} scenario, where the detection probabilities are completely reversed. 
Note that the reversed  
out-in scenario is not identical to a 
in-out scenario (nor vice versa) due to the asymmetry discussed above.

\subsection{Configurations and benchmarks}
\label{sec:benchmarks}

We evaluate our approach for different sizes of the experience memory $H$. We also experiment with the two options for resetting the experience memory discussed in Section~\ref{sec:memory}, referred to as reset1 and reset2. Furthermore, as already mentioned, we run tests using three different regression algorithms to estimate the detection event probabilities,  linear~\cite{linear}, decision tree~\cite{tree} and Bayesian~\cite{bayesian}.

We compare the proposed approach with three different policies:
(i)~\textit{Wait:} The drone acts in a conservative way and always waits for the computation to complete before moving to the next point of interest. 
(ii)~\textit{Go:} The drone acts in an optimistic way and always moves to the next point of interest right after sensing completes.  
(iii)~\textit{Random:} The decision whether to wait or to go is taken by flipping a coin.  
In addition, as the absolute reference, we use  an \textit{oracle} policy, 
which takes decisions based on Equation~\ref{eq:bestdecision} but 
has perfect knowledge of the true underlying event detection probability. 
Note that none of these benchmarks require an experience memory. 

To compare the different configurations and policies 
under the same conditions, the (random) generation of detection events for each of the test scenarios 
(following the respective probabilities) is recorded in a trace file, which is replayed during the execution of the mission. More specifically, each trace includes the presence or absence of a detection event at each point of interest and time of day (with a granularity of $1$ hour). 
We randomly generate $20$ such traces. The results presented in the sequel are the averages over all the respective mission executions.
Unless stated otherwise 
$procT$ is set to $10$ seconds, which is the measured average execution time 
of the computation on the RPi companion board of our drone.

\begin{figure*}
    \centering
    \includegraphics[width=0.85\linewidth]
    {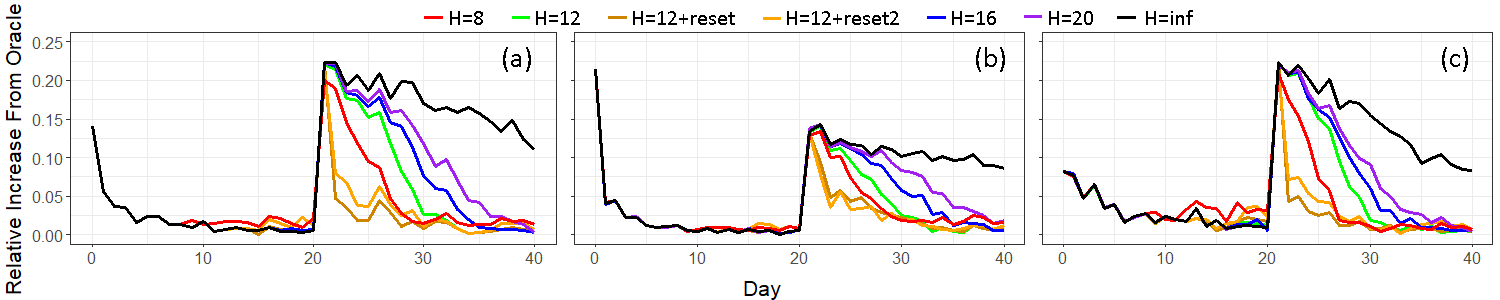}
    \caption{Experience memory comparison. Changing world. (a)  
    out scenario, (b) 
    out-in scenario, (c)  
    in-out scenario.}
    \label{fig:past_chng}
\end{figure*}

\subsection{Different experience memory sizes in a stable world}
\label{sec:stable}

In a first set of experiments, we study how the proposed approach performs in a stable world, 
where the drone performs the same mission every day for $30$ days. On the very first day (day $0$), where the drone has no past experience, it decides according to the Wait policy. We test different sizes of the experience memory $H$=8,12,16,20. We also run experiments for $H$=$\infty$, where the drone never forgets any of the past experience. In all cases, we use the decision tree regression method. 
The results are presented in Figure~\ref{fig:past_fixed} for the three different mission scenarios 
(out, out-in, in-out). Each line plots the relative increase (RI) of the mission time vs the mission time achieved by the oracle (lower is better). 

We observe that most memory-limited configurations perform very close to the one with the infinite experience memory, converging to the performance achieved by the oracle by day $12$. The results achieved for $H$=8 are visibly worse, indicating that this size is insufficient, More precisely, the RI  
for $H$=8,12,16,20 is $1.8\%$, $0.7\%$, $0.7\%$ and $0.6\%$, respectively, averaged over all mission scenarios, 

Since $H \geq$ 16 does not improve the results significantly, we set $H$=12 
and perform  
experiments for each of the reset options (reset1 and reset2). The results are also shown in Figure~\ref{fig:past_fixed}. 
An interesting observation is that the reset2 method is  
less stable than all other $H$=12 configurations, leading to worse decisions. This is because this method resets the experience memory more aggressively, forcing the drone to learn from scratch even though this is not really needed. The reset1 method, on the other hand, implements a more conservative anomaly detection heuristic, avoiding unnecessary resets of the experience memory. More specifically, reset1 has a false positive rate (FPR) of $0.1\%$, whereas for reset2 this is $3.1\%$. As a result, the RI of $H$=12 with reset1 and reset2 after day $12$ is $0.7\%$ vs $1.3\%$ for the  
out scenario, $0.3\%$ vs $0.7\%$ for  
out-in, 
and $1.0\%$ vs $1.7\%$ for the  
in-out scenario. Note that resetting the experience memory does not improve performance vs the simple $H$=12 configuration. This is expected given that the world is stable, but as will be shown below resetting becomes important when the world changes.

\subsection{Different experience memory sizes in a changing world}
\label{sec:change}

Next, we explore the performance of the proposed approach for a world that changes and all probabilities are reversed (see Section~\ref{sec:baseplan}). The change occurs on day $21$ 
and the experiment goes on for another $20$ days to  
observe the adaptation. 
Looking at the results, shown in Figure~\ref{fig:past_chng}, we see that smaller experience memories improve adaptation, while infinite experience memory 
is problematic. 
This is because a smaller experience memory makes it easier to forget old experiences. 
Also, both $H$=12 configurations with the reset logic 
detect the change and adapt faster than $H$=8, 
by learning from scratch ( 
reset1 slightly outperforms reset2 in most scenarios due to the reasons discussed above). 

The adaptation profile of all variants is similar 
in all cases. However,  
in the 
out and in-out scenarios, right after the change, all variants perform worse than day 0 (where the drone adopts the Wait policy) as the previous experience is completely invalid.  
The impact of wrong decisions is not as grave in the  
out-in scenario. The reason is that before the change the drone has learned to go in the majority of waypoints, and 
after the change the Go policy 
performs 
less suboptimal than  
Wait  
before the change (see Section~\ref{sec:policies}).  

\subsection{Different probability estimation algorithms}
\label{sec:estimation} 

\begin{figure}
    \centering
    \includegraphics[width=0.80\linewidth]{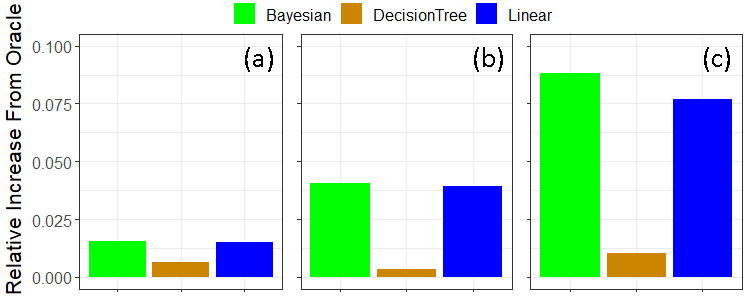}
    \caption{Probability estimation algorithms in the stable world. (a)  
    out scenario, (b)  
    out-in scenario, (c)  
    in-out scenario.}
    \label{fig:prob_prediction}
\end{figure}

\begin{figure*}
    \centering
    \includegraphics[width=0.85\linewidth]{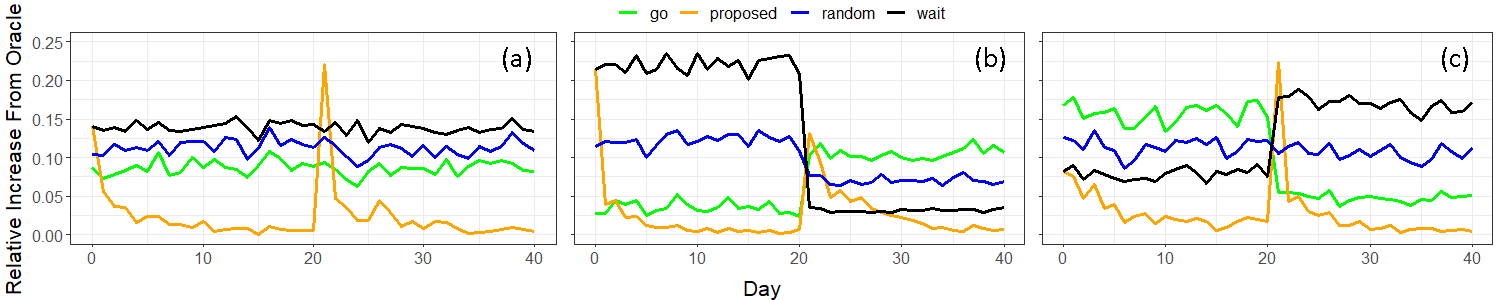}
    \caption{Policy comparison. Changing world. (a)  
    out scenario, (b)  
    out-in scenario, (c) 
    in-out scenario.}
    \label{fig:policy_chng}
\end{figure*}

Figure~\ref{fig:prob_prediction} shows the results achieved by the $H=12$-reset1 variant in a stable world, when employing different regression algorithms for the probability estimation (Section~\ref{sec:benchmarks}). 
The decision tree  
performs better than the Bayesian and linear regression method in all mission scenarios, with an average RI to the oracle of $0.7\%$, $4.8\%$ and $4.4\%$, respectively. Thus we keep the decision tree method in the rest of our study.

\subsection{Comparison with other policies in a changing world}
\label{sec:policies}

In the 
following, we compare the $H$=12-reset1 configuration of the proposed approach vs the benchmark policies introduced in Section~\ref{sec:benchmarks}. 
Figure~\ref{fig:policy_chng} shows the results for a world that changes  
as in Section~\ref{sec:change}. 

Until day $20$, when the world remains stable, the proposed approach  
outperforms all benchmarks for all mission scenarios. 
Note that in the  
out scenario the Go policy consistently outperforms Wait. The reason is that, even though there as as many waypoints with event detection probability $0.1$ as with $0.6$, there is a bias in favor of Go (the distance from $0.1$ to $0.5$ is larger than from $0.5$ to $0.6$).   
This difference becomes  
bigger in the  
out-in scenario as 
the majority of the waypoints visited by the drone have  
event generation probability $0.1$ hence the optimism of the Go policy pays-off even more. The situation is reversed in the  
in-out scenario, where Wait outperforms Go during the first $20$ days as most waypoints  
have event detection probability $0.6$. As expected, the Random policy performs  
between Go and Wait in all scenarios.

We now focus on the second phase of these experiments, after the world changes on day $21$.  
In the  
out scenario, 
all benchmarks continue to perform 
as before the change occurred. This is because 
the total number of visited waypoints with $0.1$ or $0.6$ event detection probability remains the same (though the probability at each waypoint is reversed). 
In contrast,  
in the  
out-in scenario, 
after the change 
Wait becomes better than Go because most waypoints visited have the ``opposite'' event detection probability $0.6$ vs $0.1$ before the change. Note however, that after the change Go does not perform as bad as Wait did before the change, due to the aforementioned bias in the event detection probabilities.
This is also why the Random policy performs closer to the oracle after the change. 
The reverse situation  
holds in the  
in-out scenario, 
where Go becomes a better policy than Wait after the change as most waypoints have merely $0.1$ event detection probability. 
Note that Fig~\ref{fig:policy_chng}c is not a mirror of Fig~\ref{fig:policy_chng}b for the reason explained in Section~\ref{sec:baseplan}.

The proposed approach clearly outperforms all other policies in all mission scenarios, also after the world changes. As discussed in Section~\ref{sec:change}, adaptation comes with a penalty at day $21$ 
because at that point the drone has invalid experience and takes wrong decisions in most waypoints. 
But this suboptimal behavior is immediately detected and the approach quickly converges to the oracle. 

\subsection{Policy comparison for different computation times}

\begin{figure}[h]
    \centering
    \includegraphics[width=0.9\linewidth]{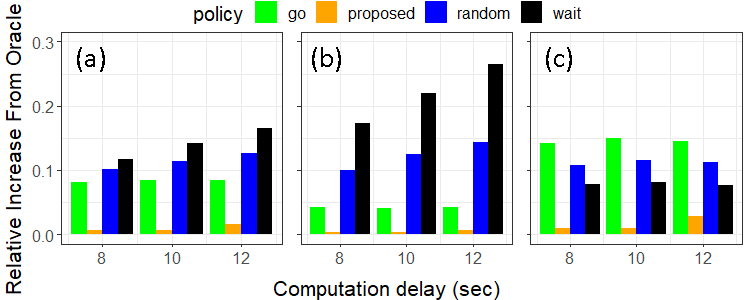}
    \caption{ 
    Different computation times in the stable world. (a)  
    out scenario, (b)  
    out-in scenario, (c)  
    in-out scenario.}
    \label{fig:comp_policy}
\end{figure}

Finally, we compare the proposed approach with the other policies for the stable world scenario, varying the computation time $procT$. Note that 
larger computation times increase the penalty of wrong decisions and the gain of correct decisions. 
Figure~\ref{fig:comp_policy} shows the average performance from day $12$ to day $30$, where our mechanism runs with full experience memory. It can be seen that the proposed approach outperforms the Wait, Go and Random policies by up to $25.8\%$, $14.0\%$ and $13.6\%$, respectively. In absolute terms, 
this saves up to $5.2$, $3.5$ and $2.8$ minutes of mission time. 

\section{Conclusion}
\label{sec:conc}

In some applications, the data acquired by the drone's sensors must be processed 
to determine 
if further action is needed at a given point of interest. We have presented an approach for learning whether to wait for the data processing to finish  
or move on to the next point of interest, 
controlling the drone accordingly. An extensive evaluation is provided for different scenarios and configuration options, also comparing with other decision policies, showing the proposed approach can handle a wide range of situations and quickly adapts to changes 
converging to  
performance  
close to that of an oracle. 

A possible extension to this work is to investigate a wider space of control decisions, where the drone could also set the flying speed when it  
decides to go. Another direction is to investigate swarm scenarios, where the area of interest is scanned by multiple drones. In this case, a drone that wrongfully decides to move to the next waypoint may be aided by another drone that is closer to the point of interest where further action is needed, in case this detour is more beneficial.

\section*{Acknowledgments}

This work has been funded in part by the Horizon Europe research and innovation programme of the European Union, under grant agreement no~101092912, project MLSysOps.

\bibliographystyle{IEEEtran}
\bibliography{bib}

\end{document}